\pdfoutput=1
\documentclass[letterpaper, 10 pt, conference]{ieeeconf}  % Comment this line out if you need a4paper

\IEEEoverridecommandlockouts                              % This command is only needed if 
\usepackage[dvipdfm]{graphicx} 
\usepackage{amsmath} % assumes amsmath package installed

\usepackage{color}
\let\labelindent\relax
\usepackage{enumitem}
\usepackage{subcaption}
\usepackage[colorlinks = true,
            linkcolor = black,
            urlcolor  = blue,
            citecolor = black,
            anchorcolor = black]{hyperref}

\title{\LARGE \bf
An Approach to Deploy Interactive Robotic Simulators on the Web for HRI Experiments: Results in Social Robot Navigation
}

\author{Nathan Tsoi, Mohamed Hussein, Olivia Fugikawa, J. D. Zhao, and Marynel V\'{a}zquez %
  \thanks{This work was supported by a 2019 Amazon Research Award. M. Hussein was supported by the National Science Foundation
Grant No. 1910565.} %
  \thanks{Artwork provided by Xaiver Ruiz, Yale College Computing and the Arts.}
  \thanks{The authors are with the Interactive Machines Group at Yale University, 51 Prospect Street, New Haven, CT 06511 and Rutgers University-Camden, 303 Cooper St, Camden, NJ 08102, USA. %
    Contact: nathan.tsoi@yale.edu} %
}
\begin{document}

\maketitle
\thispagestyle{empty}
\pagestyle{empty}

\begin{abstract}

Evaluation of social robot navigation inherently requires human input due to its qualitative nature. Motivated by the need to scale human evaluation, we propose a general method for deploying interactive, rich-client robotic simulations on the web. Prior approaches implement specific web-compatible simulators or provide tools to build a simulator for a specific study. Instead, our approach builds on standard Linux tools to share a graphical desktop with remote users. We leverage these tools to deploy simulators on the web that would typically be constrained to desktop computing environments.  As an example implementation of our approach, we introduce the SEAN Experimental Platform (SEAN-EP). With SEAN-EP, remote users can virtually interact with a mobile robot in the Social Environment for Autonomous Navigation, without installing any software on their computer or needing specialized hardware. We validated that SEAN-EP could quickly scale the collection of human feedback and its usability through an online survey. In addition, we compared human feedback from participants that interacted with a robot using SEAN-EP with feedback obtained  through  a  more  traditional  video  survey. Our results suggest that human perceptions of robots may differ based on whether they interact with the robots in simulation or observe them in videos. Also, they suggest that people perceive the surveys with interactive simulations as less mentally demanding than video surveys.

\end{abstract}

\IEEEpeerreviewmaketitle

\section{Introduction}
\begin{figure*}[bt!p]
    \includegraphics[width=\linewidth]{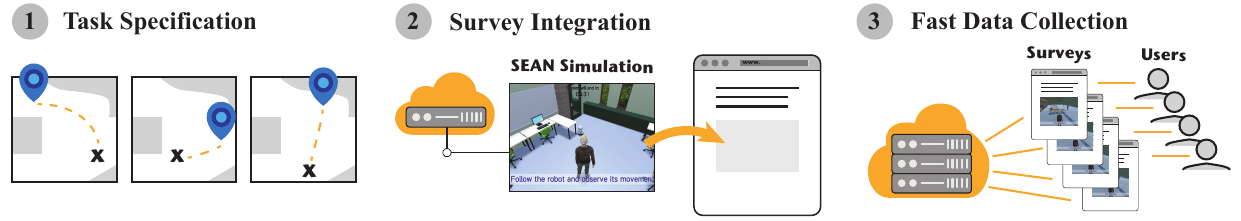}
    \caption{With SEAN-EP, researchers can scale HRI experiments in the context of navigation via 3 steps: (1) experimenters specify navigation tasks in the simulator, (2) they integrate interactive simulations based on the tasks with online surveys, and (3) they collect data in parallel from multiple users. See the text for more details.}
    \label{fig:method-flow}
    \vspace{-0.5em}
\end{figure*}

Mobile robot navigation with and around people, i.e. \textit{social navigation}, is important for robotic applications in human environments, including service robotics \cite{thrun1999minerva, evers2014development,triebel2016spencer}, healthcare \cite{calderita2020designing}, and education \cite{ahlgren2013socially,kanda2007two}.
%ahlgren2013socially
Thus, social robot navigation has long been studied from a technical and experimental perspective \cite{kirby2010social, lu2014layered, trautman2015robot, mavrogiannis2019effects, kruse2013human}. Yet, there is no agreed-upon protocol for evaluating these systems because: (a) robots often have different capabilities, making comparisons difficult; (b) implementing robust navigation systems is hard, thus baselines do not necessarily represent the state of the art; and (c) there is a lack of standard human-driven evaluation metrics because the context of navigation tasks can significantly alter what matters to users \cite{luber2012socially, torta2013design}. These issues have hindered advancements  and make it difficult to understand the key challenges that the social robot navigation community faces today.

While simulations may not be perfect replicas of the real world, we believe that they provide a viable path towards standardizing the evaluation of social robot navigation systems in Human-Robot Interaction (HRI). Our rationale is twofold. First, simulations have long been leveraged to conduct early tests, stress tests, and verification of robotic systems \cite{nourbakhsh2005human, park2013simulation, steinfeld2009oz}. They provide controlled environments to systematically study critical application scenarios, and can be used for benchmarking  \cite{madaan2020airsim, aihabitat, aws, crowdbot}. Second, simulations can be integrated with real-time robotic software, facilitating sim-to-real transfer. For example, interfaces for the Robot Operating System (ROS) have enabled running robot stacks in a variety of simulators  \cite{takaya2016simulation, airsim2017fsr, hussein2018ros, talbot2020benchbot}. 

Given prior work in robotics simulations, what is the key challenge that prevents us from fully leveraging simulations for evaluating systems in HRI? The problem is that the evaluation requires human input because the social aspects of robot navigation are subjective in nature. One option for gathering human input is to utilize web-based surveys and crowd workers. However, modern simulations are compute-intensive applications designed for local use in a desktop computing environment. That is, these simulators are \textit{rich-client} applications that provide rich functionality independent of a remote server %(and are often subject to complex software dependencies) 
-- in contrast to thin-client applications which are heavily dependent on remote processing, like browser-based web applications. % with complex software dependencies provided by the desktop environment. 
This makes the modern simulators inaccessible to crowd participants who are limited to browser-based web applications.

Making a rich-client application, like a robotics simulation, available on the web is a non-trivial task. Perhaps one could think of re-implementing the software under the constraints of a web browser \cite{taivalsaari2008web} and application-specific or web server-specific modules such as \cite{chen2016bringing}. However, some re-implementations are too complex, time consuming, or even infeasible due to the lack of specific dependencies such as a programming language, physics engine, or rendering engine. Another option could be to use specific solutions that make applications such as word processor programs available in a web browser \cite{chen2016bringing}. Unfortunately, these solutions do not generalize well to  robotic simulators that require high performance graphics rendering via specialized hardware and libraries such as OpenGL.
%
%which are composed of two components: a server and a client. The client's dependencies are met by a user's web browser, avoiding the need for complex installations and application specific hardware. The web server provides interfaces for data retrieval and persistence. Making a rich-client desktop applications available on the web is a non-trival task that has been accomplished in a variety of ways such as re-implementing software under the constraints of a web browser \cite{taivalsaari2008web} and application-specific or web server-specific modules such as \cite{chen2016bringing}. Often it is technically infeasible or too time intensive to re-implement a rich-client simulation system to run in a web browser and specific solutions that make applications such as word processor programs available in a web browser do not generalize well to systems such as robotic simulators that require high performance graphics rendering via specialized libraries such as OpenGL.
%
These challenges %associated with making typical rich-client simulations available on the web 
have often restricted human evaluation of social robot navigation via crowd-sourcing to video surveys (e.g., \cite{pokle2019deep, BalajeeWilliasm21}). While videos may lead to comparable results to in-person studies in some cases \cite{woods2006methodological}, they are passive mediums with low interactivity \cite{xu2015methodological}. 

In this work, we propose a method of making rich-client, interactive robotic simulators accessible at scale on the web. As an example implementation, we introduce the SEAN Experimental Platform (SEAN-EP), an open-source system that allows roboticists to gather human feedback for social robot navigation via online simulations, as illustrated in Figure \ref{fig:method-flow}.
Though our implementation uses the Social Environment for Autonomous Navigation (SEAN) \cite{Tsoi_2020_HAI} as the underlying simulator, any rich-client simulator that runs in Linux could be deployed using our method. % because our approach builds on standard cloud infrastructure and modern web technologies. % through which remote users can access the rich-client simulation in a standard web browser. In our implementation, users can control the motion of a human avatar and interact with a virtual robot controlled through ROS in the backend.
%Heavy computation is delegated to cloud servers such that users do not need specialized hardware to interact with the robot.

We validated our implementation and its usability through an online study about social robot navigation. Further, we %compared the human feedback about a social robot that we got from this study with feedback gathered by other participants who watched videos of  the same human-robot interactions experienced in the simulations. Our results suggest that human perceptions of robots may differ between interactive simulation and non-interactive videos. Participants perceived the interactive survey as less mentally demanding.
investigated whether human perceptions of robots differ based on whether they experience human-robot interactions via online simulations or watch them through  videos. Interestingly, our results suggest that interactive surveys are less mentally demanding than non-interactive video surveys.

In summary, our work has four main contributions:

\begin{enumerate}[align=left,leftmargin=0em,labelsep=0.2em,itemsep=0em,parsep=0.5em]
    \item A novel approach to deploy rich-client robot simulation environments at scale using standard web technologies. This method allows one to quickly gather human feedback in HRI.
    \item SEAN-EP, a specific  instantiation of the proposed approach based on the Social Environment for Autonomous Navigation. SEAN-EP is open-source and available at  \href{https://github.com/yale-sean/social\_sim\_web}{https://github.com/yale-sean/social\_sim\_web}.
    \item Validation of our example implementation (SEAN-EP) through an online study about social robot navigation.
    \item An experimental comparison of interactive simulations and videos for studying human perception of robot navigation.
\end{enumerate}

\section{Related Work}
\label{sec:related_Work}

\subsection{Robotics Simulation Environments}
\label{ssec:robsim}
Progress has been made on developing  photorealistic simulations which bridge the gap between virtual worlds and reality \cite{xiazamirhe2018gibsonenv, isaacURL}. With their high visual fidelity and responsiveness, game engines such as Unity %\cite{unity} 
and Unreal Engine %\cite{unreal} 
have proved indispensable to robotic simulation of flying and mobile robots \cite{airsim2017fsr, hussein2018ros,konrad2019simulation, guerra2019flightgoggles}. Several of these simulation environments integrate with ROS to achieve realistic robot control and transfer results to the real world. % \cite{airsim2017fsr, guerra2019flightgoggles}.

Within social robot navigation, crowd simulation and modeling of pedestrian behavior have improved as well \cite{tai2018socially,curtis2016menge, aroor2017}. However, there has been less work on combining crowd models with robotics simulation for robot navigation in human environments. One exception is the Social Environment for Autonomous Navigation (SEAN) \cite{Tsoi_2020_HAI}, which we leverage in our work. SEAN builds on Unity and integrates with ROS, making it a good option in terms of photorealism and future sim-to-real transfer.

Modern robotics simulations are rich-client applications meant to run in a desktop computer or powerful gaming laptop. While some simulators may utilize frameworks that provide the option to compile to WebGL for deployment on the web, like Unity, there are many challenges and limitations to this approach. This includes lack of direct access to IP sockets from WebGL due to security implications, limitations in rendering and illumination, lack of threading in JavaScript, and limited access to hardware \cite{bakri2016virtual}. 
%
%Because rich-client applications depend on many capabilities not present in a web browser environment, it is challenging or impossible to port these systems to run in a web browser via WebGL.
While specific limitations can be addressed via engineering workarounds on a case-by-case basis, this approach %requires changing the simulator to deal with constraints specific to the WebGL environment and lacks the speed and 
lacks the flexibility of our method.
Our method % is %able to quickly deploy rich-client simulation environments without modification.
%designed to 
works with any kind of rich-client simulation that runs in Linux, % and uses an OpenGL-based rendering engine, 
even if it requires interacting with other software such as ROS components. 
%In cases where robot control via ROS is required, running purely in a web browser is not possible due to the system requirements of ROS.

\subsection{Leveraging the Web in HRI}

%Online access to existing simulators are similar in that the dependencies of these systems typically make them difficult to access remotely. 
%Prior works have developed web-compatible simulations based on Unity and WebGL. For example the Unity Game engine has been used to develop a web based simulation to test modalities for conveying instructions during evacuation scenarios \cite{robinette2014assessment}.
%\cite{nayyar2020exploring, xu2020much}
%ROS applications can be made accessible on the web via RMS, but to the best of our knowledge there is no existing body of work focused on making existing rich-client robotic simulators available to users on the web.

While traditional HRI experiments are conducted in person, prior work has explored faster mechanisms that leverage the accessibility of the Web. For example, \cite{chernova2010crowdsourcing} explored using a two-player online game to build a data corpora for HRI research. Also, \cite{robinette2014assessment, nayyar2020exploring, xu2020much} developed web-compatible simulations based on Unity and WebGL, although these systems have the limitations discussed in Sec. \ref{ssec:robsim}.  

Especially during early HRI system development, it has become common practice  to gather human feedback via online video surveys \cite{zlotowski2012navigating,takayama2011expressing,dragan2013legibility,pokle2019deep,BalajeeWilliasm21}.
%rueben2017framing,
Research has indicated that there is a moderate to high level of agreement for subjects' preferences between live and video HRI trials \cite{woods2006methodological}. Our work contributes to a better understanding of experimental methods  by comparing  human feedback from a video survey with feedback from an interactive survey.
 
Our approach is inspired by internet frameworks that provide methods to remotely interact with robotic systems and simulators. For example, the Robot Management System (RMS) \cite{toris2014robot} and Robot Web Tools \cite{toris2015robot} provide software for building web-based HRI interfaces and demonstrate their approach in a variety of tasks, including remote robot operation. % and building cloud-based knowledge bases from robot experiences. 
These tools allow web clients to interface with ROS and the Gazebo simulator %using browser-based clients. This is achieved  
by transforming ROS-specific data streams into formats compatible with a web browser. While ROS applications can be made accessible on the web via RMS, our approach is relevant to all kinds of rich-client simulators that run on Linux, not just Gazebo. %,  regardless if they connect to ROS processes.
%without requiring any changes to the system, such as re-implementing portions within a web-compatible framework.
%{\color{red}say something that separates this work from our work...}

Lastly, RoboTurk \cite{mandlekar2018roboturk} allows for rapid crowdsourcing of high-quality demonstrations for robot learning in the context of manipulation. %This was initially achieved through real-time teleoperation of a robot in simulation and later demonstrated on real robot arms. 
While our method could be used in the future to gather data for learning navigation policies through teleoperation, in this work we explore giving users control of a human avatar in the simulation. This methodology aims to bring their online, virtual experience closer to real-world human-robot interactions.

\section{Method}
\label{sec:method}

\begin{figure}[tb!p]
\begin{center}
    \includegraphics[width=.95\linewidth]{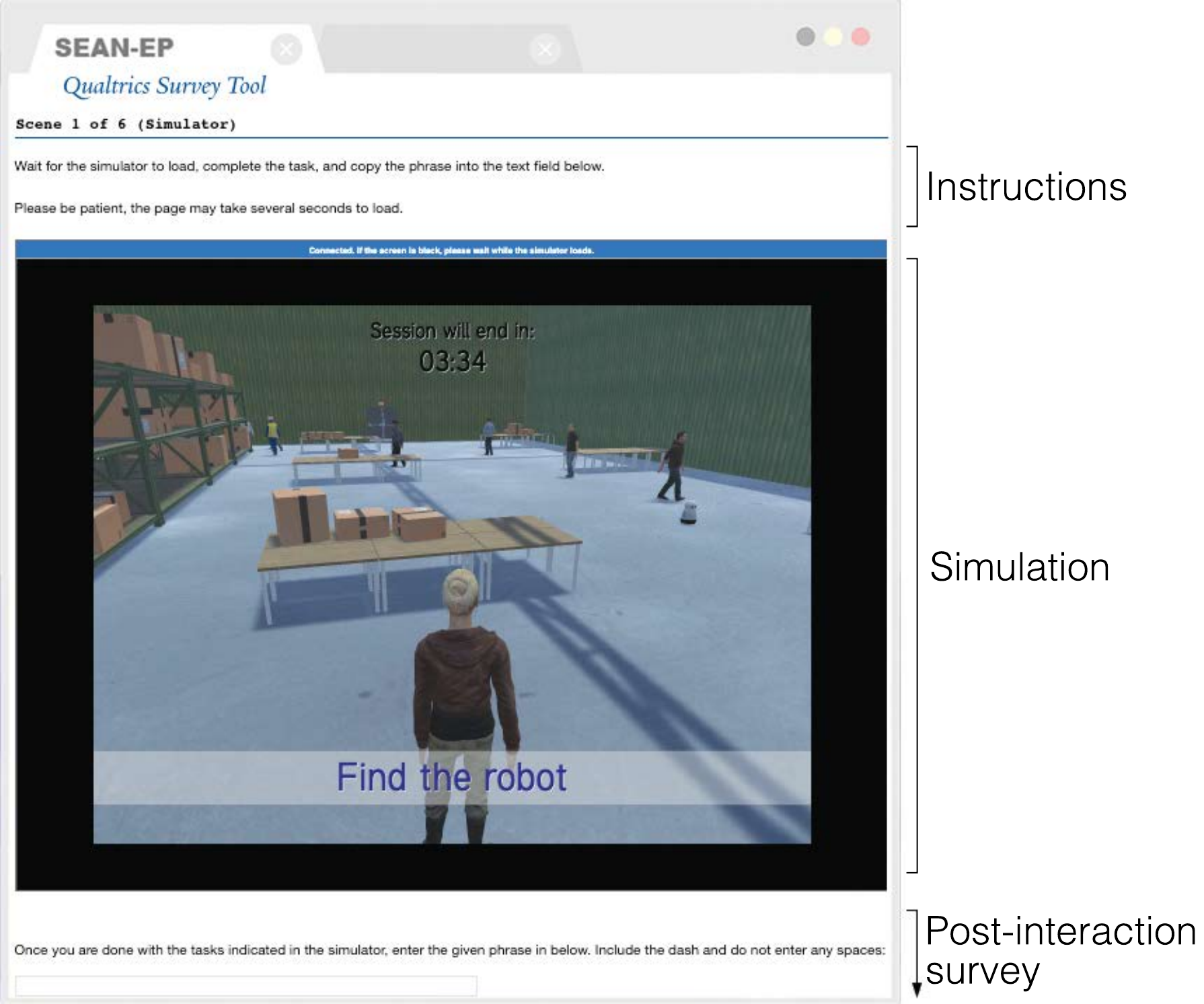}
    \caption{Screenshot of Qualtrics survey with embedded SEAN simulation. Best viewed in digital form. }
    \label{fig:example-survey}
\end{center}
\vspace{-1em}
\end{figure}

We propose a general approach to deploy the graphical user interface (GUI) of rich-client robotic simulators on the web to facilitate and scale HRI  experiments. %Standard web browsers can be used to interact with rich-client robot simulators without changes to the simulator. 
Our approach builds on standard tools for graphical desktop sharing in Linux.
%It requires no adaptation for cross compilation to other targets such as WebGL.
It does not require the adaptation of rich-client simulators to other technologies, such as WebGL. %, 
%to embed the simulator's graphical user interface (GUI) into online surveys and crowd-sourcing tools. %, such as Prolific \cite{prolific} or Amazon Mechanical Turk. 

%Figure \ref{fig:method-flow} illustrates the process through which 
%With our proposed approach, one can conduct online HRI experiments where each participant interacts with their own instance of a simulation.
%{\color{red}As opposed to in-person studies which are usually conducted with participants sequentially, %during a web-based study, up to the per-study maximum number of participants may complete the study in parallel.
%our method enables each participant to interact with their own instance of the simulation in parallel during an experiment.
%This is achieved as follows:

%{\color{red}NEW PARAGRAPH}
With our approach, researchers can create interactive HRI surveys. These are surveys that, in some parts, include simulations in which the participant interacts with a robot.
Figure \ref{fig:example-survey} shows an example online survey used to study social robot navigation. This simulation was embedded in the survey using a particular instantiation of our approach, as later described in Sec. \ref{sec:seanep}. After the simulations end, the online survey can query the participant for explicit feedback about his or her experience interacting in the virtual world.

Importantly, our method addresses parallelization challenges inherent to online studies typically run via crowdsourcing platforms.
While simulation systems for in-person HRI studies are usually designed for one participant at a time, online surveys must cope with a potentially large number of people who participate in the study simultaneously. Our method provides a mechanism to scale simulations designed for a single user to many users in parallel. This is possible without changes to the underlying system. 

%With our approach, researchers can create interactive HRI surveys. These are surveys that, in some parts, include simulations in which the participant interacts with a robot. For example, Fig. \ref{fig:example-survey} shows an example page of an online survey used to study social robot navigation. This simulation was embedded in the survey using a particular instantiation of our approach, as later described Sec. \ref{sec:seanep}. After the simulations end, the online survey can query the participant for explicit feedback about his or her experience interacting in the virtual world. 

%We demonstrate  scaling HRI studies with our approach in Sec. \ref{sec:system_eval}. 
The next sections describe in detail our method to make rich-client robotics simulations accessible on the web. We  evaluate an implementation of this approach in Section \ref{sec:system_eval}.

\subsection{Making Interactive Simulations Accessible on the Web}
\label{ssec:scaling}

We propose to make rich-client simulations available in a standard web browser by running them on a remote server, and using a Virtual Network Computing (VNC) server to share the GUI of the simulator with a remote user. %In particular, because modern robotics simulators have graphics-intensive requirements, we suggest to use a TurboVNC server for performant data transfer along with VirtualGL. However, other VNC servers could be used if OpenGL was not required by the simulator.

While remote users typically connect to a VNC server via a desktop VNC client running on their machine, we use a  browser-based VNC client running on the host server to allow browser-based access to the simulator GUI. The browser-based VNC client renders the GUI on a web page, through which our system can accept user input for the simulator, e.g., keyboard commands. 

%use noVNC, an open source VNC client, with websockify \cite{martin2015novnc} to accept user input via a regular browser and render images from the desktop on that same browser. To the best of our knowledge, this is the only VNC client that supports VirtualGL and input-output via a web browser.

One important consideration when exposing the GUI of a simulator on the web as described before is that users are unauthenticated and untrusted. Thus, it is important that the GUI of the simulator does not provide mechanisms to launch other processes on the remote server.

Because VNC connections are designed to be used by a single user, we propose the use of a web-based process orchestration tool to deploy and manage a large number of concurrent simulation environments. We call this tool the ``Process Manager'' because it controls the execution of processes associated with each simultaneous user. This tool is further described in the next section, where we explain in detail how to scale data collection on a single host.

% allows many concurrent users to simultaneously interact with robots via the simulator. Our method encompasses two methods of scaling these concurrent sessions which build upon each other: (a) scaling on a single host and (b) scaling ``horizontally'' across multiple hosts.

% There is however a limitation with any VNC server, inhibiting researchers ability to quickly deploy a program at the scale necessary for fast data collection on the web. Each VNC server instance supports only a single user at a time. Each VNC session requires loading of the necessary rich-client binaries to start and run the simulator before a user can interact in their browser. Therefore a key part of our method is a web-based process orchestration tool to deploy and manage a large number of concurrent simulation environments. This allows many concurrent users to simultaneously interact with robots via the simulator. Our method encompasses two methods of scaling these concurrent sessions which build upon each other: (a) scaling on a single host and (b) scaling ``horizontally'' across multiple hosts.

%Our orchestration system leverages Linux shell environments, Docker containers, and potentially multiple host machines to scale data collection:

\subsection{Scaling on a Single Host}
\label{ssec:single-host}

\begin{figure}[tb!p]
\begin{subfigure}[t]{\linewidth}
\centering
% SOURCE: https://docs.google.com/drawings/d/15MH3DGM-Z6EoXPLQbTNwHMk4qfMEs6zmvqbqpRcn8vk/edit
\includegraphics[width=.82\linewidth]{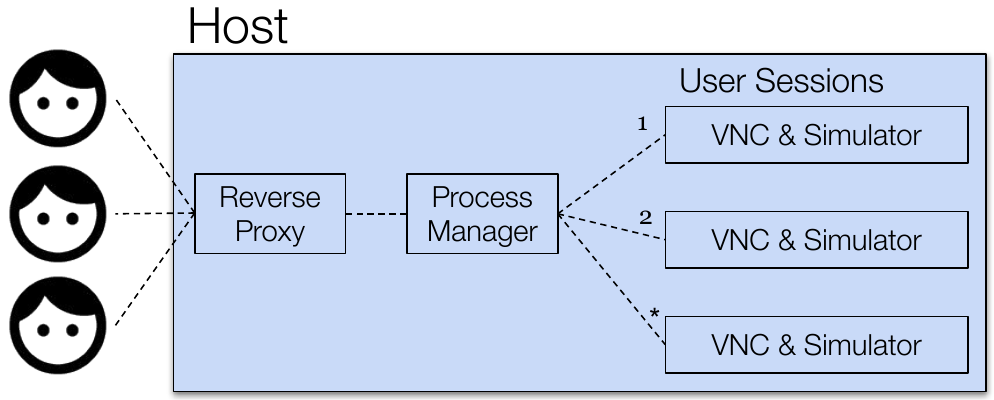}
\caption{Our method running on a single machine. The host machine accepts inbound connections via the NGINX reverse proxy.}
\label{fig:scaling_single_computer}
\end{subfigure}\\ 
\begin{subfigure}[t]{\linewidth}
\centering
\vspace{1em}
% SOURCE: https://docs.google.com/drawings/d/1IwwmFyUyZWbsTrZLP4t_Sm8UAhjfkfODhJeZxUooBOA/edit
\includegraphics[width=.65\linewidth]{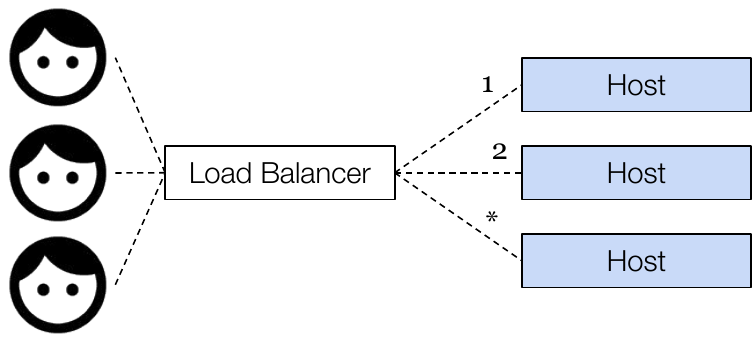}
\caption{Our method running on multiple machines. Each host  has all the components shown in  Fig.\ref{fig:scaling_single_computer}.}
\label{fig:scaling_many_computers}
\vspace{1em}
\end{subfigure}
\caption{Proposed methods to render the GUI of rich-client simulations on the web and scale HRI data collection.
}
\label{fig:scaling}
\end{figure}

To scale human feedback collection, we can run multiple instances of the simulator on the remote server and provide individual remote users access to one of them. Figure \ref{fig:scaling_single_computer} illustrates how we achieve this goal using a reverse proxy server %called NGINX \cite{reese2008nginx} 
and a Process Manager. The Process Manager is responsible for managing \textit{user sessions}, which include an instance of the simulator (including its GUI), all other components necessary for the simulator to run, and a VNC server and browser-based client for the given user. % 

The reverse proxy routes web requests that are received via Hypertext Transfer Protocol Secure (HTTPS) to specific web servers on the host machine based on the URL path of the request. The target web server may be the Process Manager, which is in charge of initiating, maintaining, and terminating user sessions, or an existing web-based VNC client within the user's session. 

Requests for simulations should have a specific URL path that includes a parameter for a unique user identifier, e.g.,  a Mechanical Turk ID. Additionally, they should include any  other parameters needed to instantiate the simulation for the user. For instance, in the example implementation described in Sec. \ref{sec:seanep}, the requests include start and goal poses for a user's avatar and a robot in the simulation. 

When the Process Manager receives a request from the reverse proxy, it evaluates if it is a new request given the URL parameters. If that is the case, then the Process Manager launches the main components that make up an interactive simulation session and quickly redirects the request to the page of the web-based VNC client that corresponds to the user. Because the VNC web page is served on the same host, the reverse proxy gets the  request that results from the URL redirection and appropriately routes it so that the user's web client can display the simulation's GUI. If an existing user requests a running simulation, the Process Manager simply redirects the request to the corresponding VNC URL. 

The Process Manager is also in charge of managing maximum session duration. Sessions are allowed to run for a configurable amount of time before being automatically shut down. Once a session is closed, the resources can be re-allocated to new sessions for other users.

Handling simulation requests as described above is beneficial in 3 key ways: (1) it is easy to integrate simulations with online surveys because a single web address (with  parameters) is used to handle all requests; (2) the entire connection between a remote user and the host machine is encrypted over an HTTPS connection; and (3) because the reverse proxy is routing requests rather than having users connect directly to each VNC instance, there is no need to expose many non-standard internet ports on the host.

Crucially, this method allows a single host to handle the multiple concurrent requests required of an online study where many users need access to interactive sessions to complete surveys.

\subsection{Scaling Across Many Machines}

Scaling of simulator sessions on a single host machine is limited by the hardware resources on the host. The ability to scale users' sessions across many machines, or scale ``horizontally,'' removes this limitation.

Horizontal scaling can be achieved by adding a Load Balancer to our proposed system. The Load Balancer receives user requests and then acts as a ``traffic cop'' to evenly distribute the requests to the available host machines, as illustrated in Figure \ref{fig:scaling_many_computers}.
This routing process must be ``session aware'' to associate users to the same host machine if they perform multiple requests.

With the guarantee that a single host will receive all requests for a unique user, the system state does not need to be shared across hosts, but can be managed in the same way as described in Sec. \ref{ssec:single-host}. Moreover new hosts can be added dynamically to handle more concurrent sessions by simply notifying the Load Balancer.

\section{SEAN-EP: A System to Scale Human Feedback for Social Robot Navigation}
\label{sec:seanep}

We created the SEAN Experimental Platform (SEAN-EP) in order to  validate our method  (Sec. \ref{sec:method}) in the context of social robot navigation. Our goal was to test our method's feasibility in a realistic usage scenario and, through this effort, verify the key tenets of scalability and usability.

SEAN-EP uses SEAN \cite{Tsoi_2020_HAI} as the core simulator. SEAN provides photorealistic virtual worlds, crowd simulations for social robot navigation, and integration with  ROS for robot control. We modified SEAN to use the Microsoft Rocketbox avatars library \cite{gonzalez2020rocketbox} for this work because these avatars are higher-quality than those  used in \cite{Tsoi_2020_HAI}.

\subsection{System Implementation}

We implemented our method as an open-source system and deployed it to virtual hosts with dedicated NVIDIA T4 GPUs using Amazon Web Services (AWS). We aimed to maximize the performance of our system and fully utilize the hardware resources to deliver a user friendly and visually appealing simulated interaction that is free of glitches or lag. To this end, we chose TurboVNC as the VNC server, which accelerates data transfer by compressing images via libjpeg-turbo. We made TurboVNC available on the web using noVNC with websockify.\footnote{\href{https://github.com/novnc/websockify}{https://github.com/novnc/websockify}} Notably, TurboVNC also supports VirtualGL for hardware accelerated 3D graphics.
%Early tests without hardware accelerated 3D graphics via VirtualGL both rendered a frame rate at below 15fps an utilized all cores on our 8 virtual-CPU host machines. With hardware acceleration, we easily achieved framerates that made the simulator appear fluid and utilized less vCPU cores.

We used the open-source NGINX server as reverse proxy and implemented the Process Manager %in the Python programming language
using the popular Flask web framework. We developed a custom configuration for NGINX to properly route requests as specified by the Process Manager. Also, we configured an AWS Application Load Balancer with sticky-sessions to make it session-aware. %, i.e., to route subsequent requests from the same user to a single host. 

The Process Manager facilitates communication between the SEAN GUI and ROS. Because each session requires its own instance of ROS, we encapsulate ROS processes in a Docker container and expose a single network port for the SEAN GUI to communicate with its ROS instance.

%Robots in SEAN are typically controlled via ROS. The SEAN GUI, based on Unity, communicates with ROS to receive robot commands and send sensor information. This requires running one instance of ROS per simulator sessions. This can be challenging due to the fact that the networking components of ROS were designed to be run only once per host machine. We overcome this challenge by encapsulating each instance of ROS in a Docker container. We then re-map and expose a single communication port to allow one instance of the SEAN GUI to communicate with its instance of ROS. The Process Manager coordinates this functionality.

We used ROS bag files as the main logging mechanism for human-robot interactions enabled by SEAN-EP. % and stored them locally on the virtual hosts. Later, we  aggregated them in shared storage in the AWS Simple Cloud Storage Service (S3) for analyses, as described in Sec. \ref{sec:system_eval}.

\subsection{Navigation Tasks}

% add background to describe the navigation tasks:
To evaluate our idea in practice and conduct an experiment about perceptions of social robot navigation (Sec. \ref{sec:system_eval}), we designed three tasks for users to complete in SEAN simulations. 
First, they had to find the robot. Second, they had to follow the robot and observe its movements, which required them to stay in proximity to the robot and observe its interactions with other people.
Third, they had to navigate to a nearby location in the environment identified with a visual landmark. This last task incentivized them to navigate around the robot to reach their destination.
Overall, these tasks 
motivated users to both interact with the robot in the virtual world and behave in naturalistic ways.

\subsection{User Interface}

%To enable users to experience the interaction with the robot in simulation in a manner similar to the way they would experience it in the real world, 
We created a new user interface in SEAN to let a user control an avatar in the simulation and make the virtual experience similar to real human-robot interactions.
We had two key requirements when designing the user interface:
%Our first requirement was to ensure that any latency in the network connection to our simulation server would have minimal impact on the person's ability to control their avatar.
it had to be accessible to a wide range of users; and it needed to be simple enough to be explained in a short introductory tutorial.
%lasting less than a minute.
Given these requirements, we chose to implement an interface that is similar to a third-person video game, albeit with simplified controls. %Users without gaming experience should be able to operate the human avatar after a short tutorial, inline with our design goals.
The main camera of the simulation follows the user's avatar as it moves. Users can press the up and down arrow keys to raise and lower the camera, changing the %providing a narrower or wider 
field of view of the environment as needed. In addition, they can use the keyboard commands W, A, S, D to move their character forward, left, backward, and right, respectively. These keyboard commands were captured in the users' browsers and seamlessly passed to the SEAN simulation using noVNC.

\subsection{Data Collection via Online Survey}

%To collect data generated in the simulator we leverage the fact that our method runs in a full desktop environment. Therefor we are able to both run ROS for robot control and collect data via ROS bags which are recorded to each host's local disk. These ROS bag files are aggregated to a single machine for later analysis. Data we collected from the simulation environment included the pose of virtual humans and the robot in the simulation, as well as images of the virtual environment. 

With our system, SEAN simulations can be integrated with standard online survey platforms via HTML iframe elements. The surveys can collect any additional data from users, such as demographic data or answers to questions about their experience in the  simulations. An example is provided in the evaluation presented in Sec. \ref{sec:system_eval}, for which we integrated SEAN simulations with a Qualtrics survey (Figure \ref{fig:example-survey}). 
%To collect human feedback for our study of human perceptions of human robot interactions and survey data such as demographic information, we leverage the Qualtrics web survey platform. 
%Because our method makes the rich-client simulation accessible through any standard web browser, it can be easily integrated with online surveys via a HTML iframe element. 

\subsection{Performance}

%Since the Process Manager is aware of the session status for each user, requests for a new users can be redirected to an interface that indicates that the simulation is starting up. The Process Manager then launch the main components that make up an interactive simulation session. This begins with the VNC server, followed by any rich-client binaries required for the simulation environment.

SEAN-EP provides users web access to interactive SEAN simulations with a small amount of load time. When a user requests a new SEAN simulation session, the Process Manager starts a complete ROS environment, the Unity-based simulator GUI, and a VNC server and client. Despite all these many programs, the start-up time for a user session is 18.9s on average. Transferring the simulator's GUI to the user's browser through noVNC takes on the order of milliseconds with a low Internet connection speed in the U.S. (e.g., on the order of 10 Mbps).
This means that the total wait time for users to access a SEAN simulation with SEAN-EP is significantly faster than compiling SEAN Unity worlds to WebGL. The reason is that the worlds are complex, resulting in simulations that are over 1.5GB in size after the compilation. 
%During this time, the user is presented with visual feedback about the status of the simulation.
%
%As a reference, both scenes from SEAN used in our system evaluation (Sec. \ref{sec:system_eval}) are over 1.5GB in size when compiled in Unity to WebGL. 
With a global average fixed broadband download speed of 77 Mbps, %\footnote{https://www.speedtest.net/global-index} 
transferring a single WebGL environment to the browser would take over 2.5 minutes.\footnote{Note that the complete SEAN simulation cannot be exported to WebGL due to ROS dependencies. Thus, we only report the expected time that it would take to load the Unity world after converting to WebGL.}

%Though SEAN cannot be run in a web browser, we compare the load time of SEAN run via our method with the time required to transfer one of SEAN's scenes to the browser. This transfer would be required if a simulator using the environment were implemented with web technologies. Both scenes from SEAN used in our experiments are over 1.5GB in size when compiled in Unity to WebGL. With a global average fixed broadband download speed of 77Mbps,\footnote{https://www.speedtest.net/global-index} transferring a single environment to the browser would take over 2.5 minutes.

\section{SEAN-EP Evaluation}
\label{sec:system_eval}

We used a Qualtrics online survey to validate the potential of our method to gather human feedback for social robot navigation. The survey included 6 interactive simulations, embedded through HTML tags, through which users could interact with a Kuri robot. The next sections detail our experimental protocol and results, with a special focus on user feedback obtained through the survey. Section \ref{sec:sim_video} later compares using this type of interactive survey versus a video survey to gather human feedback about robot navigation. The protocols for these studies were approved by our local Institutional Review Board.

\subsection{Method}
The Qualtrics survey was designed to gather feedback about robot navigation in two simulated indoor environments. One environment was a warehouse that included 15 virtual humans, a Kuri robot, and the user's avatar (Fig. \ref{fig:example-survey}). The other environment was a computer laboratory, which included one virtual human besides the robot and the user's avatar (Fig. \ref{fig:method-flow}). The goal of the user in the simulations was to first find the robot, then follow it for 30 seconds, and finally navigate to a destination identified by a visual landmark.

\begin{description}[align=left,leftmargin=0em,labelsep=0.2em,font=\textbf,itemsep=0em,parsep=0.3em]
\item[Experimental Protocol.] The survey began with a demographics section. Then, the participants were asked to behave politely in the simulator and were introduced to the task with a short simulation in the lab environment. This simulation served as a  tutorial to explain the commands that the participants could use to move their avatar, identify Kuri, and practice navigation tasks. After the tutorial, the participants experienced 6 simulations in randomized order: 3 in the warehouse environment and 3 in the laboratory. For each simulation, there were specific start and goal locations for all agents. In particular, the navigation goals of the robot and the human avatar were opposite to each other, so that they would easily encounter one another at some point in the virtual world. After each interactive simulation, the participants were asked a few questions about their experience, including whether they were able to identify the robot and whether it moved in the environment. At the end of the survey, the participants were asked about their overall experience.

\item[Robot Control.] The Kuri virtual robot was modeled after the real platform manufactured by Mayfield Robotics. It had a differential drive base, and used a simulated 2D LIDAR and odometry information from Unity to localize against a known map. All path planning and execution was completed by the ROS Navigation Stack, which used a global and local costmap for object avoidance and social navigation around virtual humans \cite{lu2014layered}. We opted to use the Navigation Stack because it is widely used by many robots, including TurtleBot platforms, PR2, and the Clearpath Husky. Also, it is used as a classical baseline and ground truth by more modern learning-based approaches \cite{pokle2019deep,Savarese-RSS-19}.

\item[Participants.] We recruited 62 participants through Prolific, a crowdsourcing platform, %\cite{prolific}
for this evaluation. Participation was limited to individuals 18 years or older, fluent in English with normal-to-corrected vision. The participants had an average age of 32 years old and 29 were female. In general, the participants were familiar with video games (M=6.1, $\sigma=1.1$), but not as familiar with robots (M=4.0, $\sigma=1.4$) based on answers on a 7-pt responding format (1 being the least familiar, 7 being the most familiar). They were paid $\$4.00$ USD for completing the survey.

\item[System Architecture.] Because our interest was testing the proposed system, we ran participants under both strategies to scale simulations (Sec. \ref{sec:method}). Half of the participants experienced simulations running on a single host machine and thus were run in small batches to avoid overloading the host. The other half interacted with simulations distributed across four machines. Virtual machines were AWS g4 EC2 instances with 32 cores, 124GB of RAM, and 15.84GB of GPU memory. Given the requirements of our simulation environment, each machine was capable of running up to 30 interactive simulation sessions in parallel, based on their GPU memory and the size of our SEAN environments.
%To ensure smooth operation regardless of system load on an individual machine
In the case of scaling across many machines, we limited the number of sessions per host below the resource-constrained maximum to 10. We also added enough hosts behind the Load Balancer to accommodate the maximum number of total simultaneous participants in our study.
%The total number of simultaneous system users is not limited by a single machine because of this horizontal scaling capability.
%Future evaluations should stress test this limit.

\end{description}

\subsection{Results}
We were able to successfully gather data by using a single host machine as well as multiple ones. With a single host, we ran on average 5 participants at a time collecting all 31 responses in about 8 hours. With the multiple host approach, we ran all 31 participants at the same time and collected all 31 responses in 1.5 hours. As a reference, the average participant took about 31 minutes to complete the survey. 

While the multiple host approach effectively reduced the time that it took to collect data through the surveys by 81\%, it required more management overhead. This included managing instances in the pool and moving collected data from the machines to a shared store for analysis.

\begin{description}[align=left,leftmargin=0em,labelsep=0.2em,font=\textbf,itemsep=0em,parsep=0.3em]

\item[Task Completion.]
%We investigated task completion in the simulations.
While not all participants followed the instructions by the book, a large majority tried and were able to complete the given tasks, validating that our system worked as intended.
In only 23 of 372 interactive trials (6.18\%) the participant's simulation session timed out before they reached their goal destination. 
Considering all 372 interactive sessions there were only 8 interactive sessions (2.15\%) in which participants did not move from the starting position. Further inspection of the data revealed that the only two sessions in which the participants failed to find the robot in the simulation corresponded to timed out sessions in which their avatar moved. These simulations were in the warehouse environment, suggesting that they tried to find the robot but the large space made it difficult for them to identify it. %Overall, these results suggest that while not all participants followed the instructions by the book, a large majority tried and were able to complete the given tasks, validating the usability of our system. 
Because the study sessions were conducted in parallel, at scale, they help confirm the scalability of our method.
%both for individual hosts and across multiple hosts.

\item[Navigation Behavior.] Using ROS logs from SEAN, we checked how often the human's avatar and the robot were close to each other based on Hall's proxemic zones \cite{hall1966hidden}. We set a threshold for intimate space of 0.45 meters, and found that in 195 sessions (52.4\%) the robot came within this distance from the participant's avatar. In terms of personal space, in 316 sessions (84.9\%) the robot came within 1.2 meters from the avatar. A benefit of simulations is that we can easily analyze proxemic behavior as shown by these results.

\item[User Experience.] At the end of the survey, the participants  reported that the survey tasks required low mental demand (M=2.26, SE=0.19) and low physical demand (M=1.61, SE=0.15) on a 7-point responding format where 1 indicated the lowest demand. They did not have to work hard to accomplish the tasks (M=2.32, SE=0.18). 

Some participants provided positive feedback for our system through open-ended questions. For example, one person said that \textit{``the game was very well made and the controls are what I'm used to with my own gaming.''} Another said they \textit{ ``found the instructions were easy to follow.''}
% source: https://docs.google.com/spreadsheets/d/1Mv4q-tByuc2VSpJEPtoNuGE-FCDXW7VI/edit#gid=1449699261
When asked if the virtual world was confusing, participants strongly indicated it was not confusing (M=2.03, SE=0.18).
%Nonetheless, 29 people expressed finding some part of the survey difficult or confusing, and 11 out of 62 participants experienced some difficulty.
However, a few participants reported confusing elements of the survey. Eight people believed the robot's motion was awkward. Additionally, 7 people thought that the control of their human avatar was unintuitive.
%Of the few users that mentioned technical difficulties, their comments involved slow loading of videos in the non-interactive condition and several users in the interactive condition refreshed the page several times when access an interactive simulation.
As a reference, the participants had an average internet speed of 105.56 Mbps (SE=11.09). 

Overall, these results validate the feasibility of our method to enable online, interactive HRI studies.

\end{description}

\section{Interactive Simulation vs. Video Feedback}
\label{sec:sim_video}

We compared the interactive human feedback obtained using SEAN-EP
%about the social robot Kuri
with feedback obtained through a video survey, which is a typical approach to online HRI studies as discussed in Section \ref{sec:related_Work}.
To this end, we recruited 62 more participants through Prolific. These participants provided feedback about the robot  based on videos of the simulations that happened as part of our prior study (Section \ref{sec:system_eval}). 

\subsection{Method}
\begin{description}[align=left,leftmargin=0em,labelsep=0.2em,font=\textbf,itemsep=0em,parsep=0.3em]
\item[Experimental Protocol.] We expanded our data from Sec. \ref{sec:system_eval} with data collected through a Qualtrics video survey. In general, the video survey followed the same format as the prior one. However, instead of having participants interact with the robot in a virtual world, each participant viewed the 6 video recordings of the simulations experienced by a participant from our validation study.
%These videos were recordings of the interactive condition.
After watching each video, they were asked about the observed robot.

\item[Hypotheses.] The data from  Sec. \ref{sec:system_eval} (Interactive condition) and the video survey (Video condition) were analyzed together to investigate two hypotheses:
\begin{description}[align=left,leftmargin=0em,labelsep=0.2em,font=\textit,itemsep=0em,parsep=0.3em]
    \item[H1.] \textit{The perception of the robot would differ between the conditions.} To test this hypothesis, we gathered ratings for the Competence and Discomfort factors of the Robotic Social Attributes Scale (ROSAS) \cite{carpinella2017robotic} (Cronbach's $\alpha$ was 0.938 and 0.746, respectively). We also gathered participants' opinions on whether the robot navigated according to social norms after each simulation session or corresponding video.
    \item[H2.] \textit{The perceived workload for  the survey in the Interactive condition would be lower than in the Video condition.} We measured perceived mental and physical demand along with effort at the end of the surveys based on responses to the following questions from the NASA Task Load Index \cite{hart1988development}: ``How mentally demanding were the tasks?'', ``How physically demanding were the tasks?'', and ``How hard did you have to work to accomplish what you had to do?''. Responses were collected on a 7 point responding format (1 being lowest, 7 being highest).
\end{description}

\item[Participants.] A total of 124 participants were considered for this experiment (62 from Section \ref{sec:system_eval} plus 62 new participants). Their average age was 34 years old and 55 of them were female. We limited participation in the same way as Sec. \ref{sec:system_eval}. Participants were paid $\$4.00$ USD for completing the survey.

\end{description}

\subsection{Results}
\begin{description}[align=left,leftmargin=0em,labelsep=0.2em,font=\textit,itemsep=0em,parsep=0.3em]

\item[Human Perception of the Robot.]
There were 2 simulation sessions out of 372 in the Interactive condition in which the participants failed to identify the robot and 53 sessions in which they said it was not moving. Meanwhile, there were 4 sessions out of 372 in the Video condition in which the participants failed to identify the robot and 51 sessions in which they said that it was not moving. We excluded these sessions from further analysis about perceptions of the robot.
% 55 people in interactive condition did not see said that the robot was not moving, 55 in video.

For the 288 pairs of sessions in which participants both saw the robot moving in the simulation and the video condition, we conducted a Wilcoxon signed-rank test for the paired data to check if the Competence and Discomfort factors from ROSAS (in 7-point responding format, 1 being lowest and 7 being highest) differed by condition. The test resulted in no significant differences for Discomfort. The median Discomfort was 2.33 for the Interactive condition and 2.17 for the Video condition.
%Average Discomfort was 2.27 for the Interactive condition (SE=0.05) and 2.30 for the Video condition (SE=0.07). 
However, the Wilcoxon test revealed significant differences for robot Competence (p$<0.0001$). The median Competence value was 3 for the Interactive condition and 3.83 for the Video condition.  Lastly, an additional paired Wilcoxon signed-rank test indicated significant differences by Condition in terms of  whether the robot navigated according to social norms. The mean rating was 3 over 7 points for the Interactive condition and 4 over 7 for the Video condition. These different results provided evidence in support of our first hypothesis (H1).

\item[Workload.]
We conducted an additional paired Wilcoxon signed-rank test to evaluate potential differences in perceived workload across Conditions. The test indicated significant differences for perceived mental demand (p$<0.01$). The median rating for mental demand in the Interactive condition was 2 points over 7, while the median for the Video condition was 3. No significant differences were found for the ratings about physical demand. The median rating was 1 -- the lowest possible -- for both conditions. Lastly, we found significant differences for how hard the participants had to work to complete the surveys (p$=0.023$). The median rating for the Interactive condition was 2 over 7 points, and the median for the Video condition was 3. These results partially support H2. 

\subsection{Discussion}

We suspect that the different results across conditions were due to the interactive nature of the simulation, which provided better opportunities for the participants to evaluate the responsiveness of the robot to their actions and to other virtual humans than the video. However, further tests are needed to validate this assumption.
%We attribute the lower mental demand and perceptions of reduced work with the Interactive survey to the simulation being more engaging than the Video survey. Greater task engagement in simulation could lead to increased engagement, decreased boredom, and reduced perceived workload.
% Our experiments compared an interactive experience for gathering human data with feedback based on videos, future works could compare these result to feedback gathered from the real world
% %, but we do not yet know how these results compare to human feedback gathered from the real-world.
% %Future work could advance understanding of these potential differences
% by comparing results between a virtual world designed to replicate a real world environment. 
For example, future tests could consider human perception of the robot and the perceived workload in both the real world and the simulated replica. Another aspect to consider is user perspective. While we used a third-person perspective in SEAN-EP, studying interactions perceived from a first-person perspective could better translate to the real world. An additional consideration for future work is the display type used by the participants.

%Our future work will study this question using one of the virtual worlds in SEAN, which was built to mimic a real laboratory at Yale. 
%Using our system, these changes require modifying only the rich-client simulation environment. No changes to our proposed method of deployment and data collection would be necessary.

\end{description}

\section{Limitations \& Future Work}

%Our method provides access to graphics-intense rich-client simulations on the web via a desktop environment. Making these environments available on the web to unauthenticated users may raise security concerns. 

%We proposed a method of making rich-client simulations available on the web. As discussed in Sec. \ref{ssec:scaling}, there are security concerns when allowing unauthenticated and untrusted users to access a system. Therefore we suggest multi-layered approach to security and limited exposure of the desktop environment as much as possible.

Our approach to make interactive simulations available on the web was effective in general. It allowed users to control their virtual avatars in rich-client simulations and quickly gather data to study social robot navigation. In the future, we would like to use SEAN-EP to also allow users to control the robot, so that we can collect example behaviors for social robot navigation. We also wish to explore other types of common navigation scenarios \cite{repiso2019adaptive, ferrer2017robot, mavrogiannis2019effects}, e.g. walking alongside a robot or  passing in narrow spaces.

%in that we tested our approach with a single robotics simulation environment, but it is applicable to all types of robotic studies in simulation and could be used in the future in many other scenarios. In future work we plan to explore the control of a robot around humans in SEAN to facilitate training of navigation algorithms. Since our system utilizes ROS for robot control, users may transition fluidly to control of a real robot. We hope to explore this further in the future.
%With the current implementation's capability, users are able to control a robot and record data via ROS in simulation. Control of a real-world robot could be made fluidly through the use of same ROS interfaces. 

We observed that most survey participants tried to complete the simulated tasks in a polite and naturalistic manner as directed. There were several people however, who explored undesired actions for their avatars. About 18\% of the participants pushed the robot in the simulation, and about 12\% collided with a human based on annotations from our video survey. In the future it is important to explore incentives for participants to reduce these undesired behaviors.

% %We envision adding other tasks to expose participants to critical robot navigation scenarios in the future. 
% In future studies we wish to explore other types of human-robot interaction scenarios such as
% %For instance, blocking the robot's path to evaluate the robot's competence in navigating conflicting social situations.
% walking alongside or in the opposite direction of the robot or passing in narrow spaces to evaluate other common navigation scenarios \cite{repiso2019adaptive, ferrer2017robot, mavrogiannis2019effects}.
% %Using our system, these changes require modifying only the rich-client simulation environment. No changes to our proposed method of deployment and data collection would be necessary.

%While we observed that in general participants who completed our interactive survey tried to complete the tasks in the simulation as indicated, there were several people who explored undesired actions for their avatars. About 18\% of the participants pushed the robot in the simulation, and about 12\% collided with a human based on annotations from our video survey. It is important to explore incentives for participants in the future to reduce these undesired behaviors. 

Lastly, we evaluated our proposed approach using a single robotics simulator. Given the flexibility of our method, we would like to see it being used to make other rich-client simulators for Linux easily available on the web. This could facilitate human feedback collection in other HRI domains. 

\section{Conclusion}
We introduced a flexible method to enable crowd-sourcing of human feedback using interactive rich-client simulators deployed on the web. We then demonstrated a particular instantiation of this approach, called SEAN-EP, in the context of social robot navigation.

We tested SEAN-EP with an online survey, which validated its ability to serve simulations to many users. Furthermore, we compared the results of evaluating robot navigation through interactive simulations using our method against evaluations based on video surveys. Our proposed interactive methodology resulted in different perception of the robot and lower mental demand for participants. 

We are excited about the potential of conducting future online HRI experiments at scale with interactive, virtual experiences. We also look forward to leveraging SEAN-EP to scale benchmarking of social robot navigation methods.
% from a human-centered perspective in the future.

% \section{Acknowledgements}

% To be added upon acceptance.

\bibliography{references}
\bibliographystyle{IEEEtran}

\end{document}